\documentclass[10 pt, conference]{ieeeconf} \IEEEoverridecommandlockouts

\usepackage{url}
\usepackage{amsmath,amssymb}
\usepackage{graphicx,subfigure}
\usepackage{color}
\usepackage{cite}
\usepackage[ruled,linesnumbered]{algorithm2e}

\usepackage{multirow}
\title{\LARGE \bf Multi-Agent Reinforcement Learning\\ for Visibility-based Persistent Monitoring}

\author{Jingxi Chen,$^{1}$ Amrish Baskaran,$^{1}$ Zhongshun Zhang, and Pratap Tokekar%
\thanks{$^{1}$Chen and Baskaran contributed equally.}
\thanks{The authors are with the University of Maryland, College Park, USA {\tt\small \{ianchen, amrishb, zszhang, tokekar\}}@umd.edu. 
This work is supported by the National Science Foundation under Grant No. 1943368 and the Office of Naval Research under Grant No. N000141812829.
}
}

\begin{document}

\maketitle
\thispagestyle{empty}

\begin{abstract} 
The Visibility-based Persistent Monitoring (VPM) problem seeks to find a set of trajectories (or controllers) for robots to persistently monitor a changing environment. Each robot has a sensor, such as a camera, with a limited field-of-view that is obstructed by obstacles in the environment. The robots may need to coordinate with each other to ensure no point in the environment is left unmonitored for long periods of time. We model the problem such that there is a penalty that accrues every time step if a point is left unmonitored. However, the dynamics of the penalty are unknown to us. We present a Multi-Agent Reinforcement Learning (MARL) algorithm for the VPM problem. Specifically, we present a Multi-Agent Graph Attention Proximal Policy Optimization (MA-G-PPO) algorithm that takes as input the local observations of all agents combined with a low resolution global map to learn a policy for each agent. The graph attention allows agents to share their information with others leading to an effective joint policy. Our main focus is to understand how effective MARL is for the VPM problem. We investigate five research questions with this broader goal. We find that MA-G-PPO is able to learn a better policy than the non-RL baseline in most cases, the effectiveness depends on agents sharing information with each other, and the policy learnt shows emergent behavior for the agents.
\end{abstract}

\section{Introduction}
Multi-robot systems are often tasked with monitoring a large environment~\cite{PM_stocastic_arrival, hari2019persistent, generalPM, pm_rob_swarms, pm_limited_range, 2d_pm}.
This includes applications such as patrolling where regularly revisiting the environment is crucial. When the robots have limited field-of-view, they may be able to see only a part of the environment at any given time. Therefore, the robots need to coordinate their actions  so that they can monitor the environment effectively. In this paper, we investigate whether a team of agents can \emph{learn} to cooperate for effective Visibility-based Persistent Monitoring (VPM).

We consider a scenario where a team of agents are tasked with persistently monitoring a 2D environment with known stationary obstacles. The agents have a limited field-of-view sensor whose visibility is obstructed by the obstacles. Any point that is not monitored accrues penalty at a rate unknown to the agents. The goal is to find trajectories for all agents so as to minimize the collective penalty.

The VPM problem is a variant of the well-studied visibility-based coverage problem. In such problems, the goal is to find a placement of robots such that they collectively see all points in the environment or maximize the area that they cover while taking into account the geometric visibility constraints. In addition to coverage, the VPM problem also requires agents to revisit the previously visited positions in order to consistently cover the whole environment. As such, the problem we study is a generalization of the coverage problem. This makes it a more challenging problem to solve, especially if the dynamics of how the penalty accrued is unknown. Likewise, VPM is a generalization of the standard Persistent Monitoring (PM) problem~\cite{hari2019persistent, SONG20141663, doi:10.1177/0278364913504011} where in order to \emph{monitor} a location, the robot needs to visit that location. The standard PM problem does not take visibility constraints into account.  

We present a deep neural network architecture termed MA-G-PPO (Multi-Agent Graph Attention Proximal Policy Optimization) for solving this problem. The network takes as input the observations for each agent and extracts a lower dimensional feature representation using Convolutional Neural Network (CNN)~\cite{cnn_features}, shares the features with other agents through Graph Attention (GAT) layer~\cite{graphAT}, and uses Proximal Policy Optimization (PPO)~\cite{schulman2017proximal} for learning the policy. While this architecture is similar to the ones that have been recently proposed for multi-agent coordination~\cite{jiang2020graph}, this is the first work to present a Multi-agent Reinforcement Learning (MARL) approach for the VPM problem. Nevertheless, our main contribution is an empirical investigation of the efficacy of MARL for VPM. 

Specifically, we investigate the following five \emph{research questions} in the context of the VPM problem: (1) Does MARL perform better than non-RL baselines for the VPM problem? (2) How does the information available to the agents affect the performance? (3) Are there emergent behaviors due to MARL? (4) How much does introducing communication improve the performance in MARL? (5) How well does MARL generalize to the number of agents?

We present our findings for each of these questions along with some conjectures that can inform future work. We hope that these findings can inform other applications of multi-agent RL to multi-robot cooperative tasks.

\section{Related work}
The stationary visibility-based coverage problem for a team of mobile agents is called Watchman Route Problem (WRP)~\cite{carlsson1999computing}, which is the mobile version of the Art Gallery Problem (AGP)~\cite{o1987art}. The goal in the WRP is to design paths for agents to minimize the time required to visually cover the entire environment. There are variants of these problems that take into account practical restrictions on mobile robots (e.g., time to stop and take measurements~\cite{tokekar2015visibility}). In general, visibility-based coverage problems are NP-hard. 


Several deterministic methods have been proposed for solving multi-agent persistent monitoring. 
\cite{REZAZADEH2019217} proposes a method of observing stochastic events at geographical nodes with a group of mobile agents by using a receding horizon sequential greedy algorithm to determine a sub-optimal policy  with a polynomial cost and guaranteed bound on optimality. This is facilitated by showing that the reward function is a monotone sub-modular set function. 
There is also existing work on 2D PM \cite{SONG20141663} that focuses on monitoring a set of target points in the environment. They proposed a planner for robots to reduce the uncertainty on all target points. While there is significant work on the PM problem, there are major differences compared to the VPM problem we study here: The environment in PM problems is typically represented as a graph with vertices or a set of points on the plane that need to be visited whereas here we have a continuous 2D environment that needs to be persistently monitored. Furthermore, in the 2D PM problems, there is no consideration of agents' visibility unlike VPM, where visiting a point may not be necessary. 

For multi-agent systems, communication between agents can be modeled as a Graph Convolution Neural Network (GCN) \cite{DBLP:journals/corr/abs-1810-09202}, in this work the agents are sharing their local observation through a graph neural network and using the attention mechanism to assign weights for aggregating the received information, finally based on this aggregated information RL is used for decision making. The GAT \cite{graphAT} is an improvement over GCN, dealing with the dynamic communication graph in multi-agent systems.
The work most closely related to ours is \cite{blumenkamp2020emergence, li2020graph}. Both papers focus on multi-robot coordination with supervised learning and reinforcement learning, respectively. However, these papers focus on simpler tasks such as navigation and static coverage. Here, we explicitly focus on VPM. Furthermore, we seek to understand the efficacy of MARL for VPM, something that has not been studied yet.


\section{Persistent Monitoring}
\label{sec:probform}
In this section, we formally describe the persistent monitoring problem. 
We represent the environment as a 2D grid world $K$. Each cell in the environment is to be monitored by the agents. We associate a reward value, i.e., negative-valued penalty, with each cell. 
Let $R^{(k)}(t)$ be the reward associated with cell $k\in K$ at time $t$. This reward is a function of the last time any agent monitored the cell $k$. 
Let $D$ be a non-negative decay rate for the reward. If the cell $k$ is within the field of view of any agent at time $t$, then the reward will be reset to 0; otherwise, the reward for cell $k$ decays with rate $D$. We define the reward associated with cell $k$ as follows: $R^{(k)}(t + 1)= 0$, if $k$ is monitored at $t$, and
$R^{(k)}(t + 1)=\max\{R^{(k)}(t) - D,  -R_{\max}\}$ otherwise.


All cells are initialized with a reward, $R^{(k)}(0) = 0$. Since $D\geq 0$, $R^{(k)}(t)$ will be non-positive. An equivalent interpretation of $R^{(k)}(t)$ is to think of it as the negative of the penalty associated with not viewing the cell $k$. The longer the time spent between successive viewings, the larger the accrued penalty. $R_{\max}$ is used to limit the penalty so that it does not grow unbounded. The objective of the persistent monitoring problem is to find an optimal policy $\pi$ to maximize the cumulative reward collected by all agents in a finite horizon of length $T$: $
\underset{{\pi}}{\text{maximize}}  \sum_{t= 0}^{T}\sum_{k\in K}^{}R_{\pi}^{(k)}(t).
$
We show the dependence on $\pi$ to represent the fact that the reward collected at any time $t$ is a function of the policy for the agents until time $t$. The standard coverage problem~\cite{GALCERAN20131258} is a special case where the $D$ is set to 0 once a cell is viewed by some agent for the first time. 


The policy maps the states to actions for each agent. Each agent can choose from one of four neighbors and staying in place. 
In general, we do not assume that $D$ is known to the algorithm; thereby making this an RL problem. 

Since the reward is a function of the history of the position of the agents, the state must contain not only the positions of the agents but also the reward values for each cell in the grid at the current time step. This ensures the Markov property. However, the state space is large and therefore, we resort to using RL to solve the problem. We investigate the effect of other types of state representation as discussed next, instead of the full grid map.

\begin{figure}[htb]
    \centering
    \includegraphics[width=0.9\linewidth]{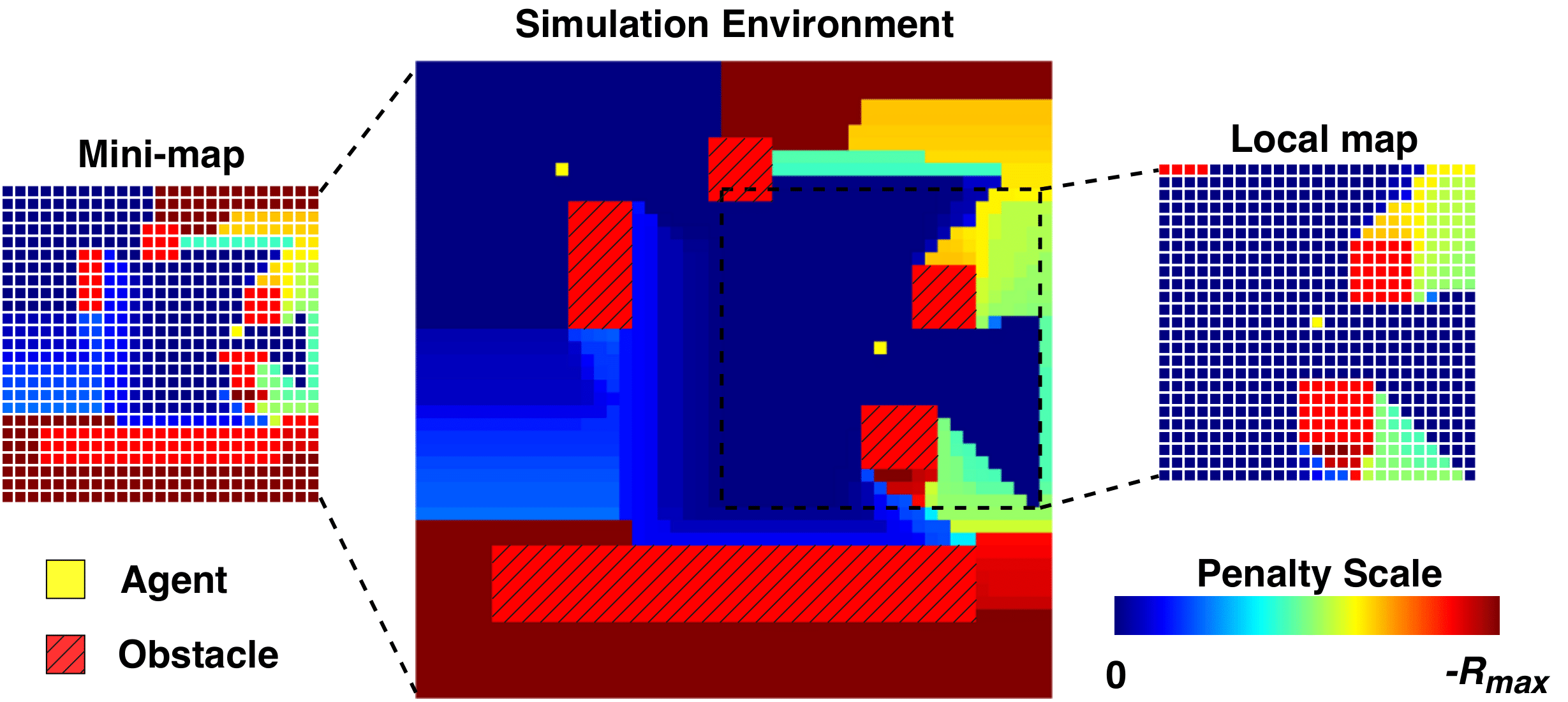}
    \caption{From left: Mini-map, Simulation environment and Local map
    \label{fig: illustration of maps}}
\end{figure}

An instance of the simulated environment is shown in Figure \ref{fig: illustration of maps}. The figure shows the following elements: (1) Agents: represented in yellow dots. (Note: in the penalty scale, the yellow region is not to represent the agents.) (2) Static Obstacles in red (diagonal fill). 
(3) Free cells: Agents can only traverse through these free cells. The color of the free cells represent $R^{(k)}(t)$ ranging from blue (0) to red ($-R_{\max}$).

We assume that each agent has a limited  visibility coverage range with a square field-of-view of side length $L$ centered at the agent. This is shown in the Figure \ref{fig: illustration of maps} within the black dashed box. We consider both scenarios where agents have access to the global information of all cells or have only the local information of the area visible at the current time step. We investigate the effect of learning with:
 (1) Mini-map: As shown in Figure \ref{fig: illustration of maps}, The mini-map is a lower resolution form of the simulation environment. The mini-map contains the position of the current agent only with penalty values and position of the obstacles. (2) Local map: As shown in Figure \ref{fig: illustration of maps}, provides a higher resolution map of the environment around the agent at the current time step (in the ego-centric frame of the agent). The local map is essentially the observation given by the sensor onboard the agent.
We investigate the effect of learning with only mini-map, with only local map, and with both local and mini-map.

\section{Proposed Approach: MA-G-PPO}
In this section, we describe the approach we use to solve the VPM problem discussed in the previous section. Our architecture uses the communication-based MARL structure~\cite{jiang2020graph} along with GAT \cite{graphAT} to model communication. We term this architecture as Multi-agent PPO with GAT (MA-G-PPO).

\begin{figure}[htb]
\centering     
{\includegraphics[height=45mm]{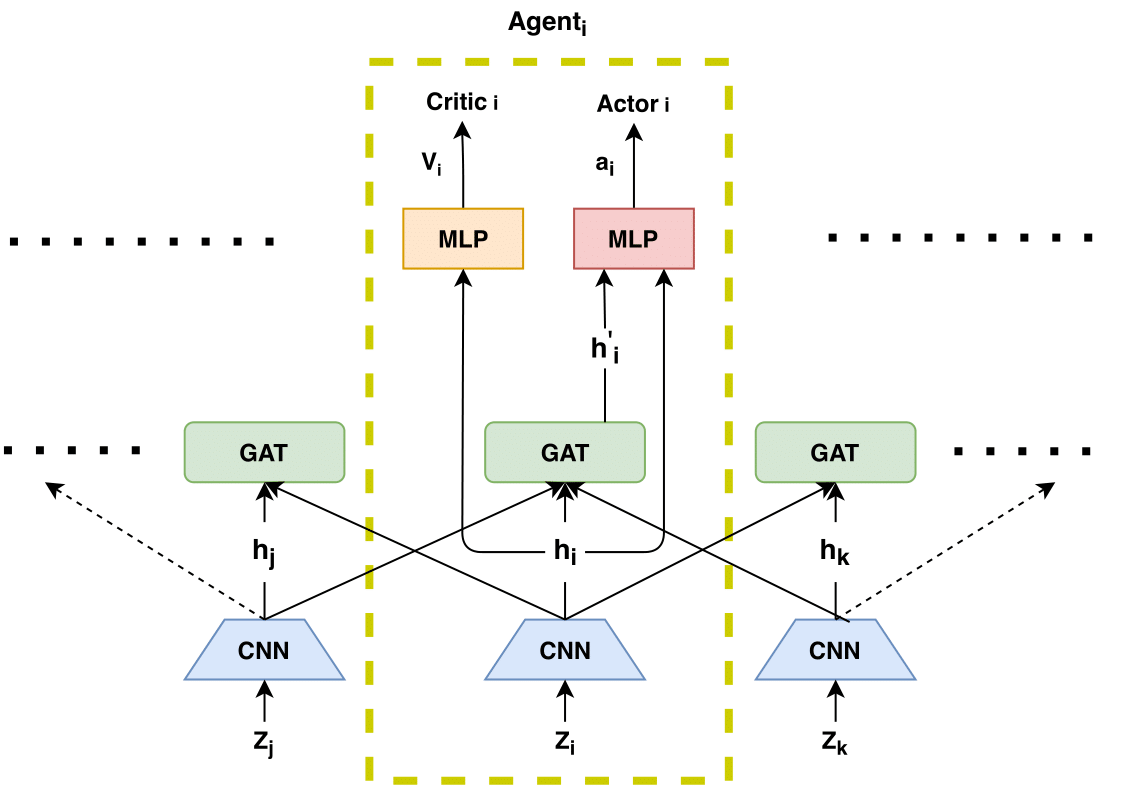}}
\caption{MA-G-PPO architecture. 
}
\label{fig:general structure}
\end{figure}

Figure \ref{fig:general structure} shows the schematic of MA-G-PPO. The structure in the yellow dashed box is replicated for all agents. Each agent gets a copy of the same structure (with identical weights). This allows us to generalize the setup by replicating this structure for additional agents. 
The structure consists of the following modules that we describe in detail: CNN for extracting features from the inputs for each agent; GAT to share information across agents; and PPO for finding the optimal policy for each agent. 


\subsection{Local Feature Extraction}
The input to the network is the observation $z_i$ for all agents $i$ at time $t$. We use a CNN to extract a feature vector $h_{i}$ from $z_{i}$. This feature vector is then passed to the GAT and shared with other agents. We investigate three types of observations $z_{i}$: (i) only the local map; (ii) only the global mini-map; and  (iii) the combined local and mini-map (as shown in Figure~\ref{fig: illustration of maps}). We represent the local map and the mini-map as an image where the cells corresponding to the obstacle, agent and free cells are 150, 200 and penalty ranging from 0 to $-R^{max}$ respectively.
In case (iii), the input $z_i$ is a two channel image consisting of the local map and mini-map images. As shown in Section \ref{sec:Experiment}, case (iii) performs the best which is not surprising since it contains more information. This embedding layer runs locally on each agent.

\subsection{Graph Attention Network}
After the local feature extraction, we use GAT in order to share the local feature vectors with neighboring agents. This is represented as a weighted edge between the agents (nodes) $i$ and $j$. Each agent aggregates all the incoming feature vectors into, $h_i'$, as shown in the Figure \ref{fig:general structure}. 

The learnable attention value $\alpha_{i,j}$ gives how much weightage agent $i$ gives to the feature vector from $j$. The attention weight value on the edge between $i$ and $j$ is calculated using: 
\begin{equation}
\alpha_{i,j}^{m}= \frac{\exp(\textsc{LeakyRelu}(\textbf{a}(\textbf{W}^{m}h_{i}, \textbf{W}^{m}h_{j})))}{\sum_{k \in N_{i}}\exp(\textsc{LeakyRelu}(\textbf{a}(\textbf{W}^{m}h_{i}, \textbf{W}^{m}h_{k})))}
\label{update cell}
\end{equation}
Here, 
$h_{i} \in \mathbb{R}^{F}$ is the feature vector of node $i$; 
$N_{i}$ is the set of neighboring nodes of node $i$; $W\in \mathbb{R}^{F^{'} \times F}$ is a learnable linear transformation matrix; $\textbf{a}: \mathbb{R}^{F^{'}} \times \mathbb{R}^{F^{'}}  \rightarrow \mathbb{R}$ is a single layer feedforward neural network which produces a real value; \textsc{LeakyRelu} is the activation function applied to the output of $\textbf{a}$. We are using multi-head attention for $m$ heads as it can capture more types of relationships and makes training stable\cite{DBLP:journals/corr/abs-1810-09202}.

After a node $i$ aggregates feature vectors from its neighbors $N_{i}$, the aggregated vector $h_{i}^{'}$ will be:
\begin{equation}
h_{i}^{'}= \frac{1}{M}\sum_{m= 1}^{M}\left(\sum_{j \in N_{i}}\alpha_{i,j}^{m}\textbf{W}^{m}h_{j}\right)
\label{update informtion}
\end{equation}
This way of updating $h_{i}^{'}$ is permutation-invariant to the order of nodes in its neighborhood $N_{i}$. This is because the attention weight $\alpha_{i,j}$ between agent $i$ and its neighbor $j$ is a function of the features $h_{i}$ and $h_{j}$.

In our simulations, we consider a fully connected network. That is $N_i$ is the set of all other agents except $i$. However, any other neighborhood relation can directly be incorporated in the architecture. For example, $N_i$ can be based on the communication range of the agents.

\subsection{Multi-agent PPO}
With the aggregated feature vector $h_i'$ from GAT along with its own feature vector $h_i$ are fed as the input to the next layers for each agent. Here, we choose PPO, a state-of-the-art actor-critic method for learning the optimal policy. Policy gradient actor-critic methods, in particular PPO, generally outperforms DQN in the context of path planning tasks \cite{pg_with_fa}. 

The PPO portion of the network consists of an actor and a critic head. The parameters of the actor are denoted by $\theta_{1}$ and those of the critic are denoted by $\theta_{2}$. Since the VPM problem is a cooperative task,  the reward for all the agent is the same. The shared reward at time $t$ is the summation of reward values for all unoccupied cells: $r(t,\theta_{1})= \sum_{k\in K}^{}R_{\pi(\theta_{1})}^{(k)}(t)$. 

Let $G(t, \theta_{1})=  \sum_{t= 0}^{T}\gamma^{t} r(t,\theta_{1})$ be the discounted sum of rewards collected from the start ($t= 0$) to the end ($t= T$) of an episode where $\gamma$ is the discount factor. Let $V^{i}(s_{t}^{i}, \theta_{2})$ be the estimate of the expected return for the state $s_{t}^{i}$ for agent $i$, $V^{i}(s_{t}^{i}, \theta_{2}) = \mathbb{E}[G(t)|s_{t}^{i}]$. This estimate is provided by the critic network. The advantage value for agent $i$ at time $t$ is: $A_{t}^{i}(\theta_{1},\theta_{2}) = G(t,\theta_{1}) - V^{i}(s_{t}^{i},\theta_{2})$. The surrogate loss in PPO for agent $i$ is: $surr^{i}(\theta_{1},\theta_{2})=\mathbb{E}_{t}[\min_{\theta_{1},\theta_{2}} \left\{ratio_{t}A_{t}^{i}, clip(ratio_{t}, 1 - \epsilon, 1 + \epsilon)A_{t}^{i} \right\}]$. Here, the $clip$ and $ratio$ are as defined in~\cite{schulman2017proximal}. We minimize the following loss function:
\begin{equation}
L(\theta_{1},\theta_{2}) = \frac{1}{S}\sum_{S}\frac{1}{N}\sum_{i = 1}^{N}(surr^{i})
\label{Loss func}
\end{equation}
where $N$ is number of agents, $S$ is the size of the mini-batch.

\section{Experimental Evaluation} \label{sec:Experiment}
We conducted a set of experiments to evaluate the proposed approach for solving the VPM problem. Our goal for the experiments was to answer the five questions posed in the introduction with the overarching goal of assessing the efficacy of MARL methods for solving the VPM problem. In the following, we first describe our experimental setup followed by a discussion of the investigation we carried out for the five research questions.

\paragraph*{Experimental Setup} We carried out simulations in a custom-built environment in Python.\footnote{The code of MA-G-PPO is available at  \url{https://github.com/raaslab/rl_multi_agent}.} The training and testing are performed on a computer with Ubuntu 16.04 LTS Operation system, an Intel(R) Xeon(R) Silver CPU, and a GeForce RTX 2080 Ti GPU. 

The MA-G-PPO network was implemented in Pytorch. 
We use the Deep Graph Library (DGL) \cite{wang2020deep} for graph neural network operations. 

The training is conducted with $T=1000$ steps per episode and testing with $T=2000$ steps per episode on $50\times 50$ maps and field-of-view $L=25$. In the following sections, we will use notation $N_{train}$ as the number of agents used during the training and $N_{test}$ as number of agents during the deployment/testing. For updating penalty value on cells, decay rate $D = 1$ and the maximum penalty value on a cell $R^{max} = 400$. The local map and the mini-map are $25\times 25$.
The size of the feature vectors are $F = F^{'} = 128$. We use three attention heads, i.e., $m = 3$. All $N_{test}$ values are averaged over 10 test runs for the learned policy. 

\paragraph*{Actual Training Time} Here is a sample training time to show the practical time constraint on the total number of training episodes we can run. For 4 agents, the typical MA-G-PPO training time for 10k episodes is almost 16 hours.

Recall that the reward values as defined in Section~\ref{sec:probform} are non-positive. 
For ease of understanding we view it as a minimization problem and look at the absolute value of the rewards i.e. minimizing total penalty.

We now discuss the results of our investigation of the five research questions. 

\subsection{Does MARL perform better than non-RL baselines for the VPM problem?}
We compare our MA-G-PPO with three non-RL baselines: Greedy Centralized Search (GCS), Travelling Salesman Problem-Cyclic (TSPC) and Random policy.

The GCS selects a destination cell for every \emph{unassigned} agent at every time instance as follows. We first find the subset of cells, say $S$, that have non-zero penalties. Then, we sort the cells in the descending order of penalties. Next, we iteratively choose the cell in $S$ with the highest penalty and add it to a candidate list, say $C$. Once a cell $c$ is added to $C$, we iterate through $S$ and remove from it any cell that is within a predefined $L2$ distance from $c$. We repeat this until $S$ is empty. Then, we greedily assign every \emph{unassigned} agent, in a predefined order, to the nearest cell in $C$ such that no cell is assigned to more than one agent. Then, the agent follows the shortest path using Dijkstra to the assigned cell. An agent is reassigned when it reaches its assigned cell or if the assigned cell is no longer present in the set $S$.



In a context where the location of nodes are determined
and the routes between then are undetermined,\cite{yu2014persistent} proposed an optimal cyclic patrolling scheme for persistent monitoring. Similar approaches have been devised for other variants of PM~\cite{zhang2019tree, zhang2021game, yu2014persistent, Baykal2020, 6630905} . We cannot directly use these strategies as a baseline since they do not take the visibility into account. Instead, we take inspiration from these methods and devise a baseline that does find a cyclic tour for all robots to follow. To find this tour, we first find points that collectively sees all points in the environment. Finding the shortest tour in an environment with obstacles is NP-complete~\cite{tokekar2016algorithms}. However, we can find an approximation of the shortest tour. In our case, we first find a set of points such that visiting these points is sufficient to ensure all points in the environment are seen at least once (we do this manually for the test environments). Then, we find a TSP tour that visits all the points using  Machine Learning Randomized Optimization and SEarch (mlrose) solver~\cite{Hayes19} 
The agents are then evenly placed along this tour and they execute the entire tour simultaneously. 

The random policy samples an action for each agent from a uniform distribution at every time instance.

\begin{figure}[htb]
\centering     
{\includegraphics[width=85mm]{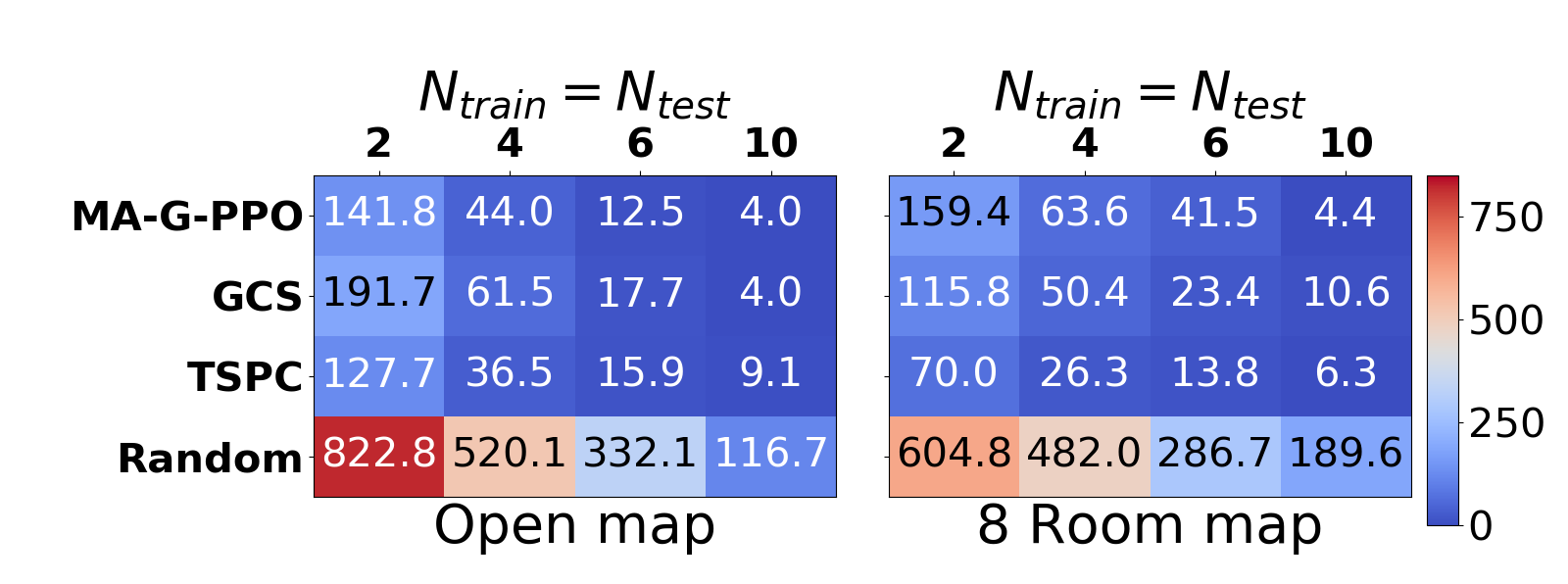}}
\caption{Comparison of cumulative penalty (unit: $10^{6}$) between four algorithms on two maps. Note that TSPC finds an approximation of the shortest tour with optimal preset points, or the near-optimal policy.}
\label{fig:comm_RL_nonRL}
\end{figure}
Figure \ref{fig:comm_RL_nonRL} shows the results of comparison between the four algorithms for a variety of $N_{train}$ and $N_{test}$ values. Not surprisingly, MA-G-PPO, GCS and TSPC always outperform the random policy. On open map, for MA-G-PPO, GCS and TSPC, as number of agents decrease from 10 to 2, the gap between their performance increases. When there are fewer agents, MA-G-PPO performs much better than GCS. But with 10 agents MA-G-PPO and GCS perform identically. This is not surprising. As number of agents increases, the solution of problem becomes closer to static coverage. TSPC follows a cyclic policy irrespective of number of agents, hence its performance degrades compared to MA-G-PPO and GCS as the solution moves towards a static coverage solution. 

\begin{figure}[htb]
\centering     
\subfigure[4 agents]
{\label{fig:4ag_policy}\includegraphics[width=27mm]{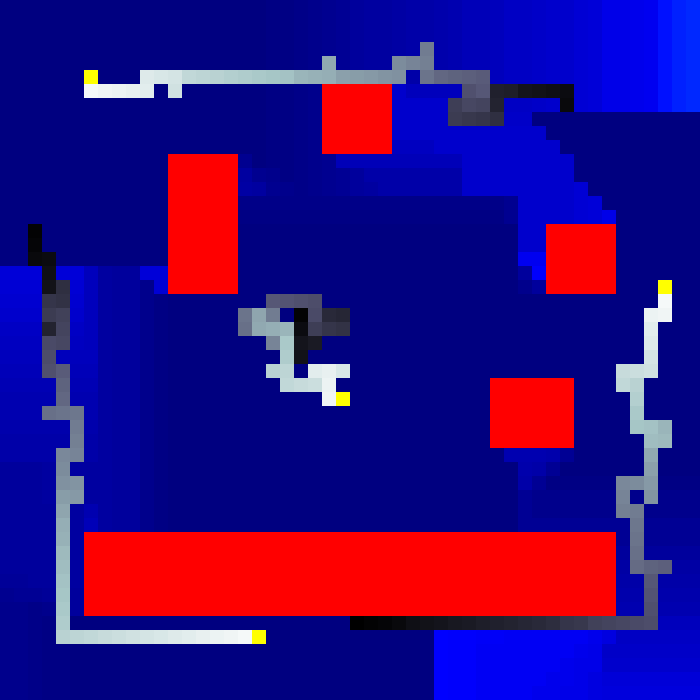}}
\subfigure[10 agents]
{\label{fig:10ag_policy}
\includegraphics[width=27mm]{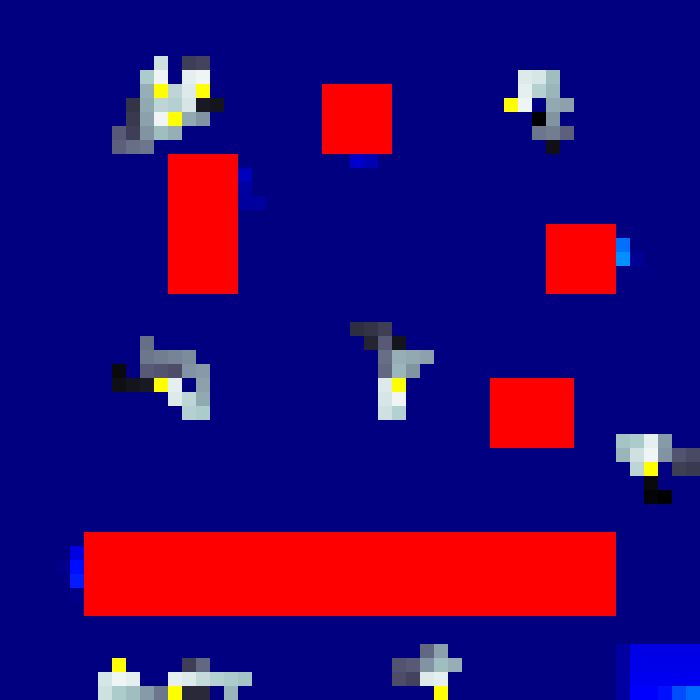}}
\caption{Learned policy figures for 4 and 10 agents in the open map.}
\label{fig: optimal policy}
\end{figure}

This can be seen in Figure \ref{fig: optimal policy}. We show the policy learned by MA-G-PPO with 4 and 10 agents on the open map. The shaded trail shows the trajectories followed by the agents. As we can see, with 4 agents, the learned policy has the agents cover larger distances effectively carrying out a tour of the environment. But with 10 agents, we see that the agents all remain in almost the same positions. When there are fewer agents, there is a need for careful planning. This is where MA-G-PPO outperforms GCS by a larger margin. Regardless of the number of agents, we see that MA-G-PPO is able to learn a good solution that outperforms GCS.

\begin{figure}[htb]
\centering     
\subfigure[4 agents]
{\label{fig:4ag_policy-8r}\includegraphics[width=27mm]{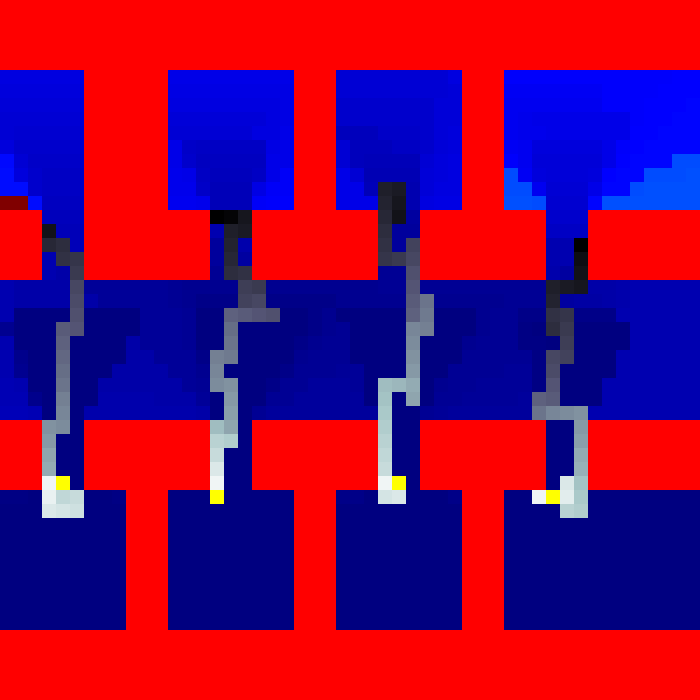}}
\subfigure[10 agents]
{\label{fig:10ag_policy-8r}
\includegraphics[width=27mm]{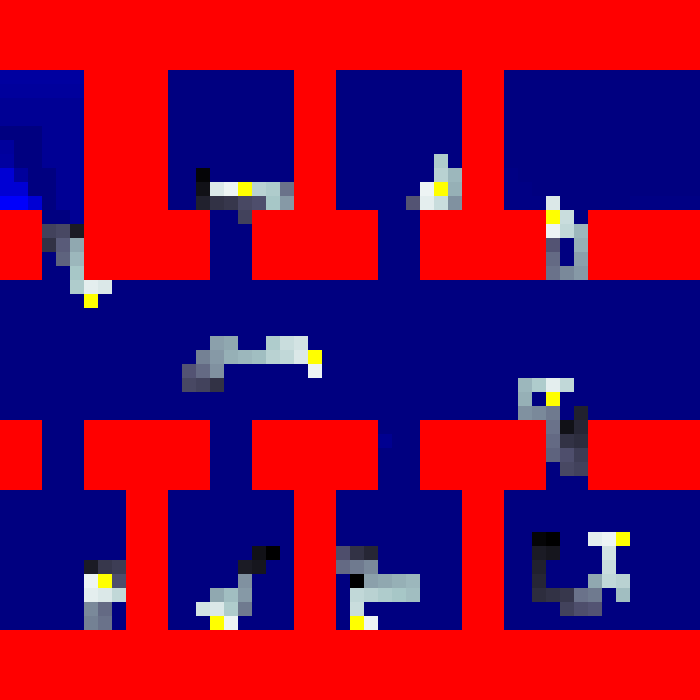}}
\caption{Learned policy  for 4 and 10 agents in the 8-Room map.}
\label{fig: optimal policy-8-room}
\end{figure}
\begin{figure} [htb]
\centering     
{\includegraphics[width=70mm]{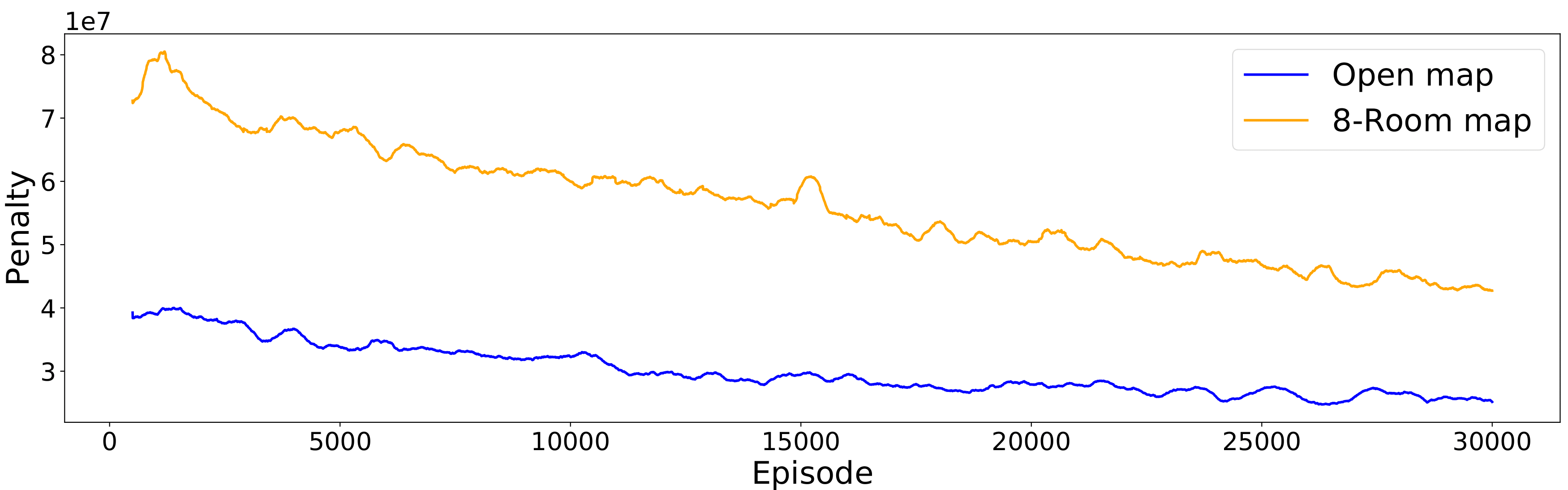}}
\caption{Rolling average of 500 episodes of the training performance for $N_{train} = 4$.}
\label{fig:8r_training_v_om}
\end{figure}

 
We also carried out the same experiment in a more structured and larger environment (Figure~\ref{fig: optimal policy-8-room}) with 8 rooms. Here, we find that GCS performs better than the policy learned by MA-G-PPO after 30k episodes. We suspect that 30k episodes are not sufficient for learning a better policy in this more structured environment. Part of the reason is that the optimal policy here requires precise movement (each agent must monitor two of the rooms, move along a vertical line, and move in a way so as to be out of phase with respect to each other). After 30k episodes, the agent learns a policy that has the first two properties but not the third. However, as we see from the trend shown in Figure~\ref{fig:8r_training_v_om} the policy has not converged after 30k episodes for the map with 8 rooms. In fact, in the open map case, the policy is learnt quickly and we see marginal improvement over the 30k episodes. But with the larger map, the policy consistently improves. We suspect that by training longer, we would end up with a policy that does outperform GCS as we see in smaller environments. In the case of a static solution, MA-G-PPO performs better than TSPC.

From the results, we conclude that MA-G-PPO effectively plans the path for the agents for VPM ranging from scenarios requiring only a nearly-static deployment to those requiring carefully chosen trajectories. We study the structure of these trajectories further in Section~\ref{sec: section C}.

\subsection{How does the information available to the agents affect the performance?}
\label{Experiment: input}
We are interested in understanding the effect of the available information on the performance of the agent. We consider three options for the input $z_i$ for each agent: the local map and mini-map independently and together.

\begin{figure}[htb]
\centering
{
\includegraphics[width=70mm]{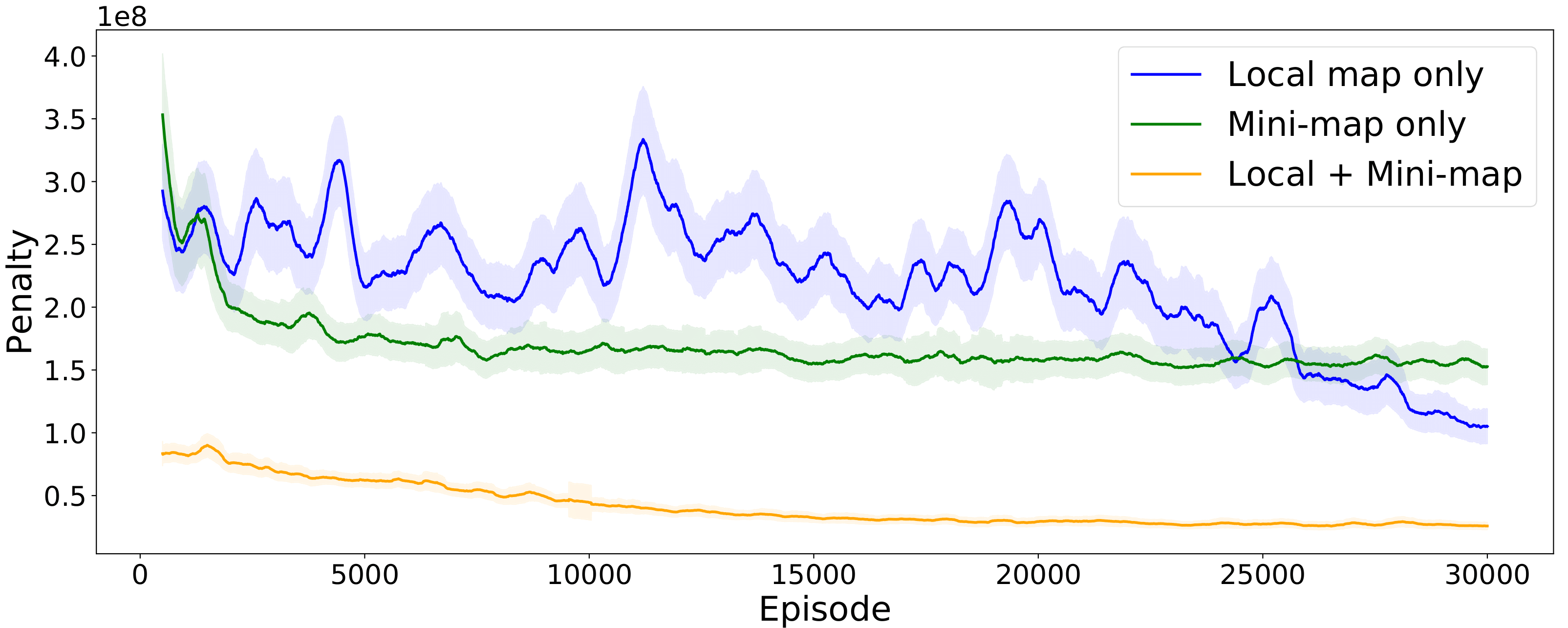}}

\caption{Training performance with three input options to MA-G-PPO using rolling average over 500 episodes and the $\pm0.5std$ error band.  }
\label{fig:baseline_comparison}
\end{figure}

We trained the policy in each of the three cases for a total of 30k episodes. We plot the total penalty accumulated across each episode for each of the inputs used, in Figure \ref{fig:baseline_comparison}. We see that training with only mini-map learns faster and is more stable than using only local map. This is not surprising since the mini-map has full access to the global information but just at a lower resolution. 
However, there is a diminishing return to training with just the mini-map. Global information at a lower resolution is not enough to plan trajectories through complex structures that do not show up on the mini-map. 

We see that the local and mini-map together far outperforms learning with just one of the prior maps, right from the outset. It must be noted however, that despite having only local map, the network is still able to perform well (even outperforming global mini-map information) given sufficient training time. Due to the high computation cost and the risk of over-fitting, a more faster and stable option is chosen. Hence, in the rest of the simulations, we use the networks trained on both local and mini-map.

\subsection{Are there emergent behaviors due to MARL?} \label{sec: section C}
\begin{figure}[htb]
\centering     
\subfigure[]
{\label{fig:4r_2ag_policy}\includegraphics[width=22mm,, height=29mm]{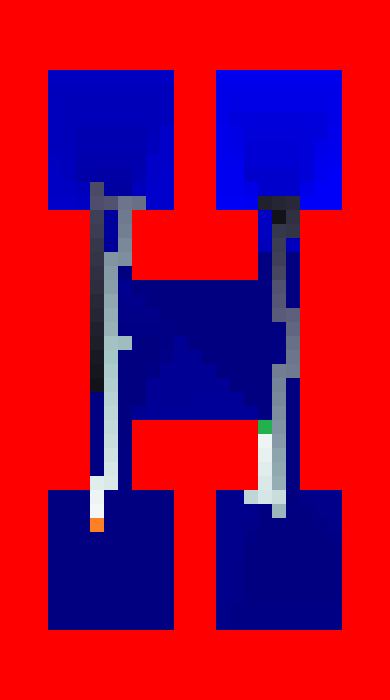}}
\subfigure[]
{\label{fig:4r_2g_phase_diff} \includegraphics[width=58mm, height = 30mm]{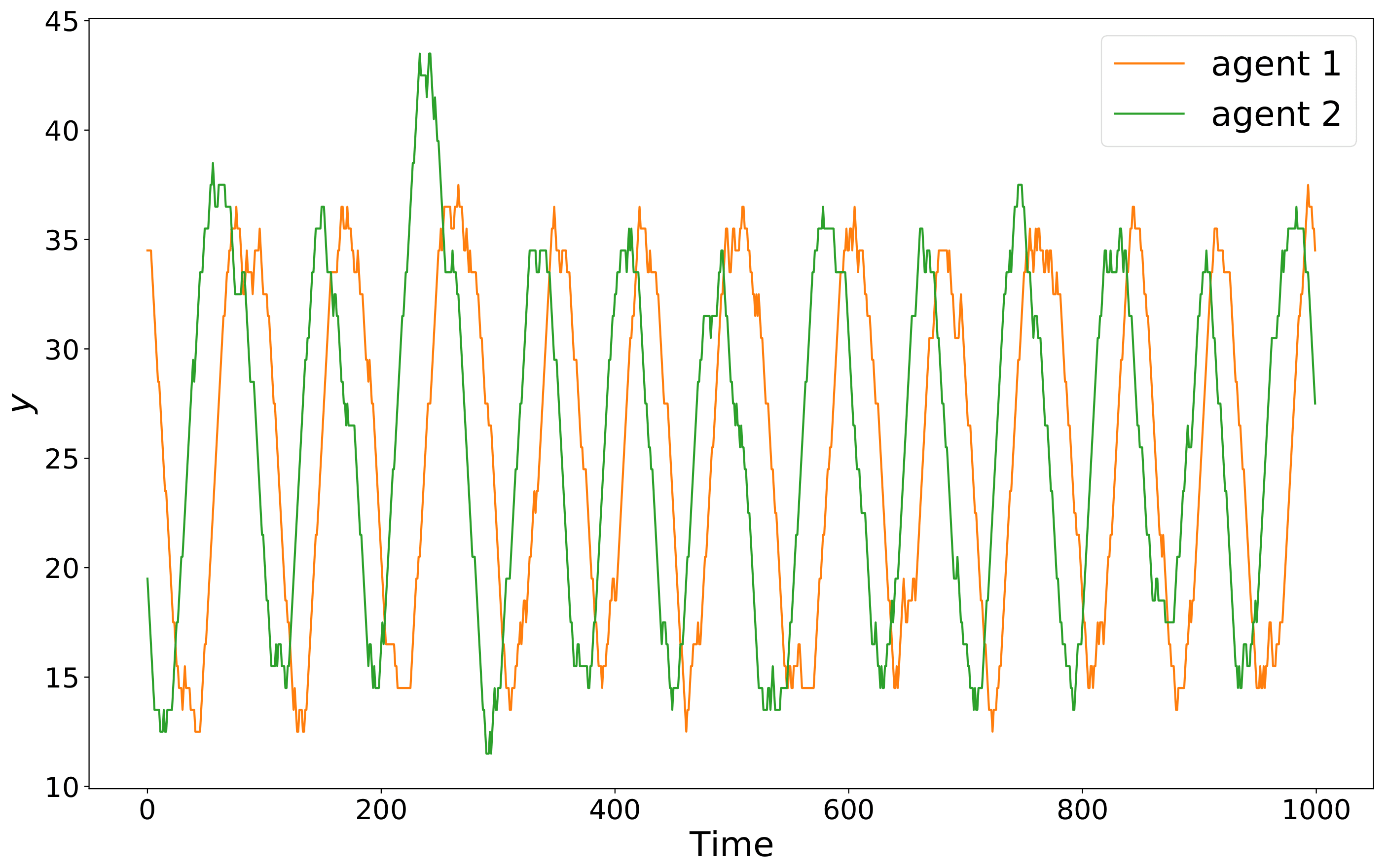}}
\caption{(a) Trajectory of learned policy for two agents in this 4-Room Map. (b) Phase difference between the agents in y-coordinates.}
\label{fig:4room_trajectory}
\end{figure}

The persistent monitoring problem has a strong underlying geometric structure. We investigated whether MA-G-PPO can find structure in the policy learned and lead to emergent behaviors. In particular, we are interested in understanding and interpreting the behaviors learned by the agent, if any. 
We investigated this through two types of environments. The first environment is structured with four rooms as shown in Figure \ref{fig:4r_2ag_policy}. There are two agents. MA-G-PPO learns a natural policy: each agent is ``assigned'' to monitor two of the rooms. Figure~\ref{fig:4r_2ag_policy} shows the trail of the agents paths. The agents oscillate between the two rooms.  Interestingly, there is a phase difference between the two oscillations. We see this in the $y$--coordinates plotted in Figure~\ref{fig:4r_2g_phase_diff} for both agents over time. We also observe that the learned policy is periodic and with a fixed phase difference between the two agents. We believe by learning a phase difference, the agents take turns monitoring the middle corridor ensuring that the penalty values there do not reach a higher value. 

\begin{figure}[htb]
\centering
\subfigure[]
{\label{fig:4ag_sine_graph}
\includegraphics[width=41mm, height = 30mm]{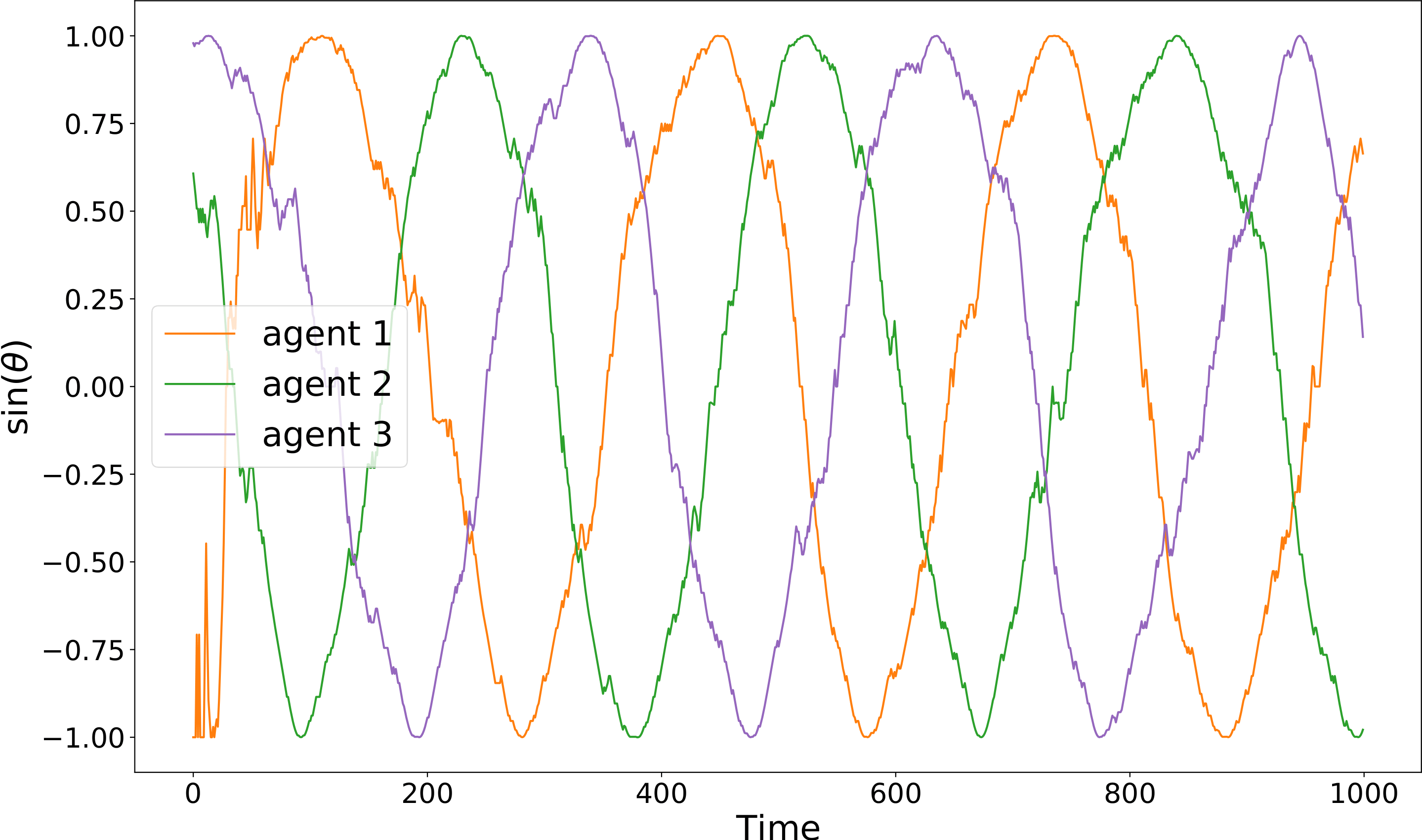}}
\subfigure[]
{\label{fig:6ag_sine_graph}
\includegraphics[width=41mm, height = 30mm]{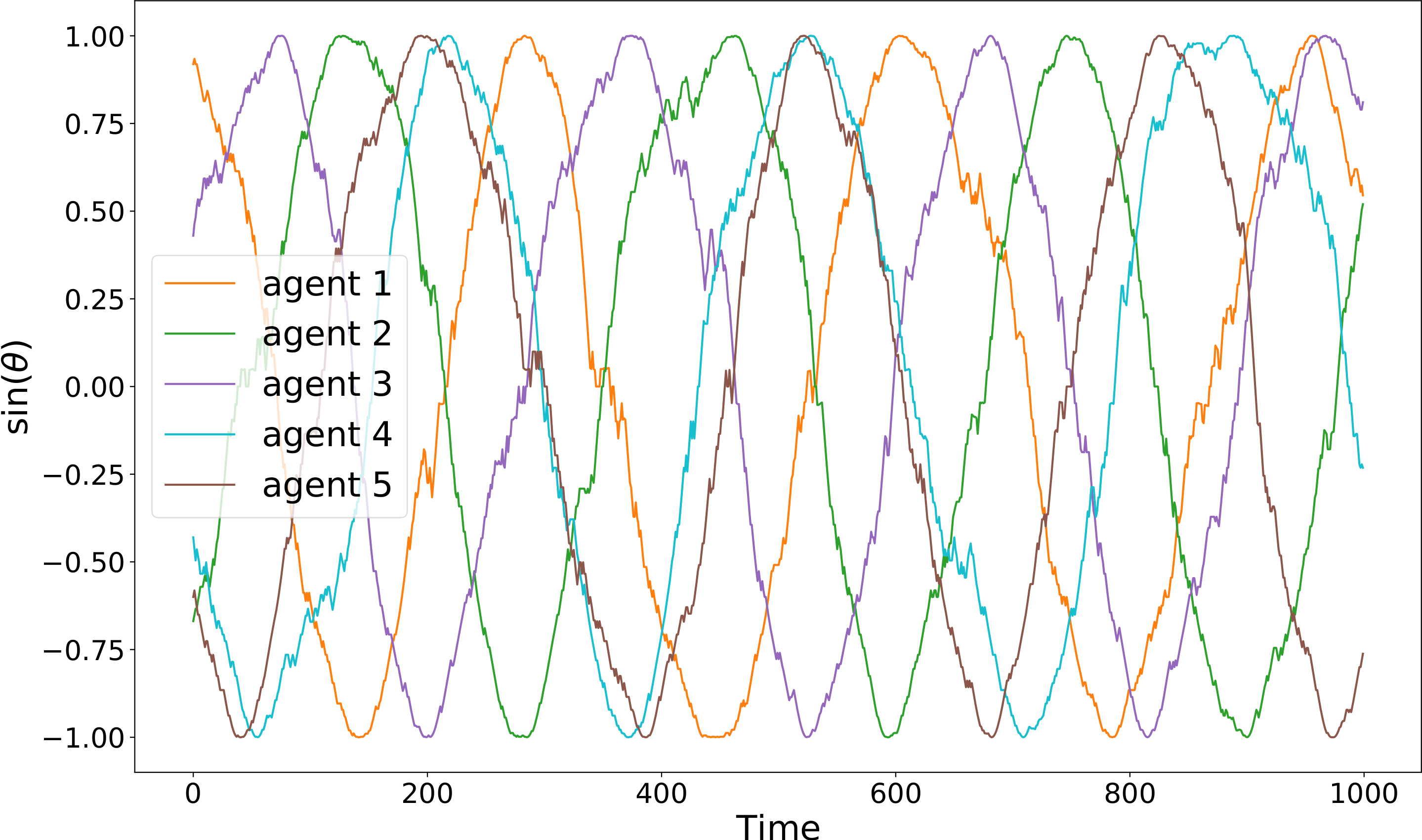}}
\caption{In open map. (a) polar-coordinate plot for $N_{test}= 4$ (c.f. Figure \ref{fig:4ag_policy}).  (b) polar-coordinate plot for $N_{test}= 6$.}
\label{fig:sin_theta}
\end{figure}

We observe a similar emergent behavior in the second type of environment which is an open one (Figure~\ref{fig:4ag_policy}). Here, we train with four agents. MA-G-PPO again learns a policy with an intuitive structure: one agent at the center remains nearly stationary throughout the episode and the other three agents move along a tour around the periphery of the environment. To observe the structure, we plot $\sin(\theta_i)$ for each of the three outer agents in Figure~\ref{fig:4ag_sine_graph}. Here $\theta_i$ are their angular coordinates with respect to the center of the environment. We can observe that the three agents follow a nearly periodic policy with a fixed phase difference as before.


In~\cite{alamdari2014persistent}, Alamdari et al. introduced the notion of \emph{kernels} for infinite horizon persistent monitoring paths. A kernel is a finite length path which when repeated forms the infinite horizon solution to the original problem. We conjecture that MA-G-PPO is able to learn the kernel in structure environments. The two sets of results shown here provide evidence of such a learned kernel (i.e., periodic paths).

\subsection{How much does introducing communication improve the performance in MARL?} \label{sec:GAT} 
To investigate this question, we compared two versions: one with communication (MA-G-PPO) and one without it (MA-PPO). The latter is the same as the former but without GAT. We train both  models with 2, 4, 6, and 10 agents on the same map. Both models are trained separately and then deployed with various number of agents.

We show the comparison between the learned policies in Figure~\ref{fig:comm_compare}. First consider the diagonal entries in the table that correspond to the case where $N_{train}= N_{test}$. We observe that in all cases, the architecture with communication outperforms the one without it. 
As the mini-map of an agent does not contain location information of other agents, it cannot predict or coordinate with each other.
GAT allows agents to communicate their global position and observation with each other, in the form of aggregated information. We conjecture that the aggregated information facilitates agents' coordination on a global scale.


\subsection{How well does MARL generalize to the number of agents?} \label{sec:gen} 
In the results discussed so far, we assume that the number of agents during training are the same as the number of agents in deployment, i.e., $N_{train}= N_{test}$. In practice, that may not be the case. For example, some robots may malfunction during deployment and retraining a policy from scratch for the new number of robots may be infeasible. Therefore, we also investigated how well does MA-G-PPO generalize when $N_{train}\neq N_{test}$. We also compare this to the generalization abilites of MA-PPO.

\begin{figure}[htb]
\centering     
{\includegraphics[width=75mm]{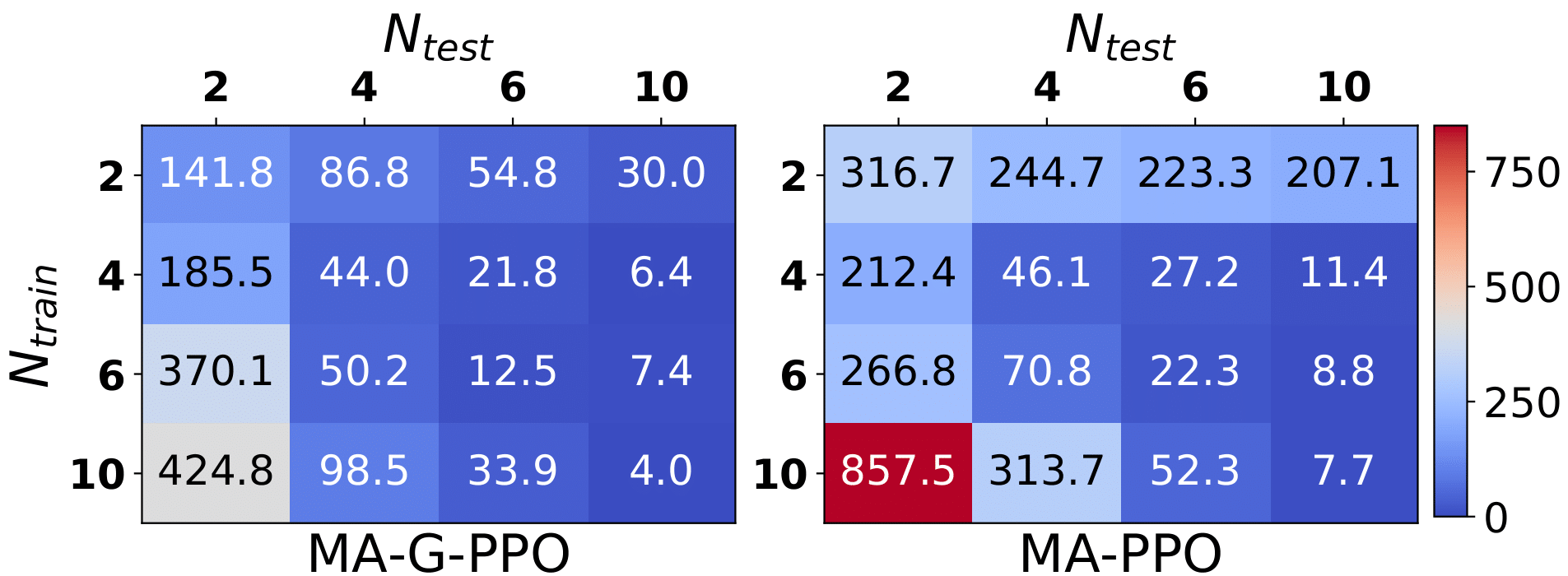}}
\caption{Average cumulative penalty (units: $10^{6}$) per episode. }
\label{fig:comm_compare}
\end{figure}

This corresponds to the off-diagonal entries in Figure~\ref{fig:comm_compare}. Consider each column in the table (corresponding to the same number of agents $N_{test}$ deployed with varying number of agents during training). Not surprisingly, the penalty is lowest when $N_{train}=N_{test}$. But we also observe that MA-G-PPO generalizes better than MA-PPO. For example, consider the second column corresponding to $N_{test}=4$. When $N_{train}=N_{test}$, both MA-G-PPO and MA-PPO perform close to each other (44.0 vs 46.1). However, as the gap between $N_{train}$ and $N_{test}$ increases, so does the gap between MA-G-PPO and MA-PPO. In the extreme case when $N_{test}=4$ and $N_{train}=10$, MA-G-PPO still yields a penalty of 98.5 compared to the much higher penalty of 313.7 with MA-PPO. A similar trend is seen in other cases.

One possible way of explaining this is that the communication help agents learn the structure of the solution (periodicity and phase difference) as described earlier. For MA-G-PPO, as we can see in Figure \ref{fig:4ag_sine_graph}, the emergent behavior learned when $N_{test}=4$ translates to the case shown in Figure~\ref{fig:6ag_sine_graph} when $N_{test}=6$. In both cases, $N_{train} =4$. We observe the same structure in the policy: one agent in the center, the remaining ones performing a tour around the periphery with the same period and fixed phase difference. Figure \ref{fig:6ag_sine_graph} shows that even if we train on four agents and deploy on six agents, the structure in the policy (one agent at the center, five along the periphery) remains similar to  the $N_{test} = 4$ case. The phase difference becomes smaller with more agents.

\begin{figure}[htb]
\centering     
\subfigure[]
{\label{fig:obs map}\includegraphics[width=30mm]{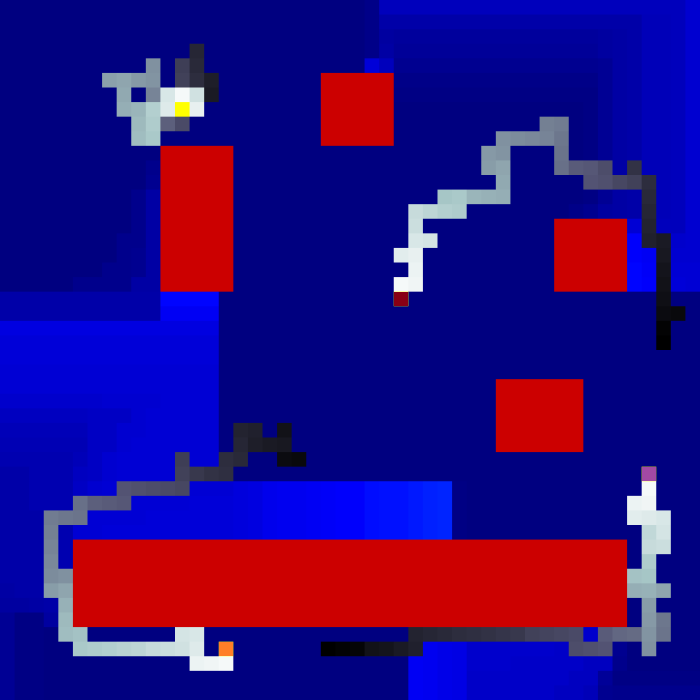}}
\subfigure[]
{\label{fig:heat map} \includegraphics[width=30mm]{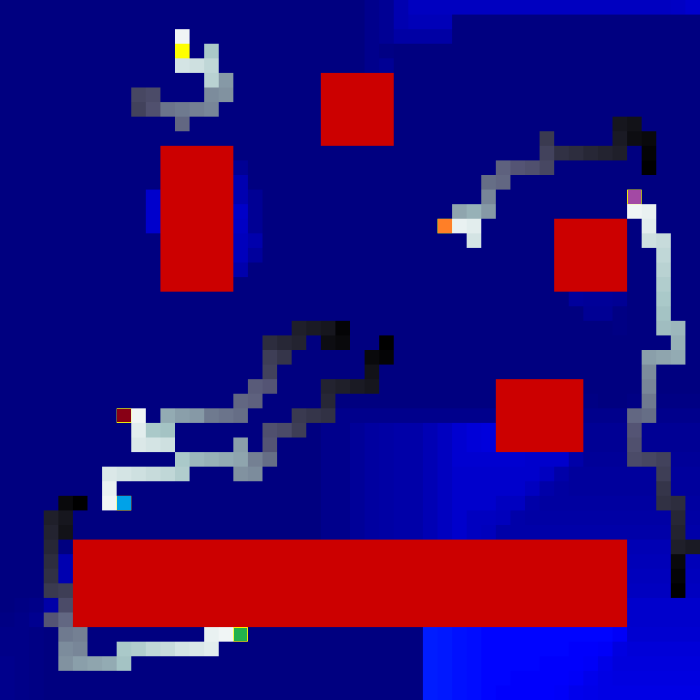}}
\caption{Training MA-PPO with $N_{train}=4$. (a) Trajectory for $N_{test} = 4$. (b) Trajectory for $N_{test} = 6$ }
\label{fig:optimal:open map_b2}
\end{figure} 

\begin{figure}[htb]
\centering
\subfigure[]
{\label{fig:baselien2_4_sine_graph}
\includegraphics[width=40mm, height = 30mm]{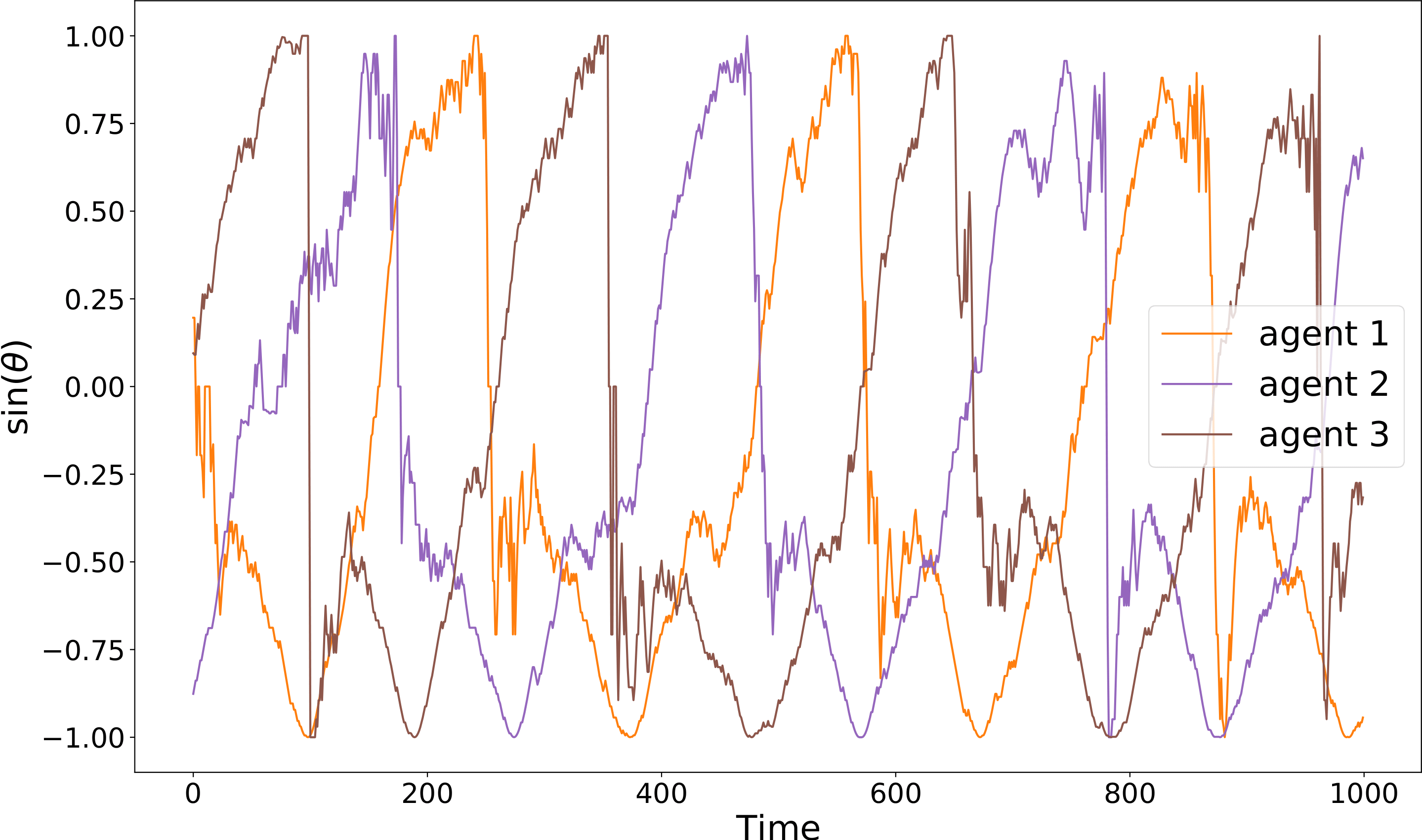}}
\subfigure[]
{\label{fig:baselien2_6_sine_graph}
\includegraphics[width=40mm, height = 30mm]{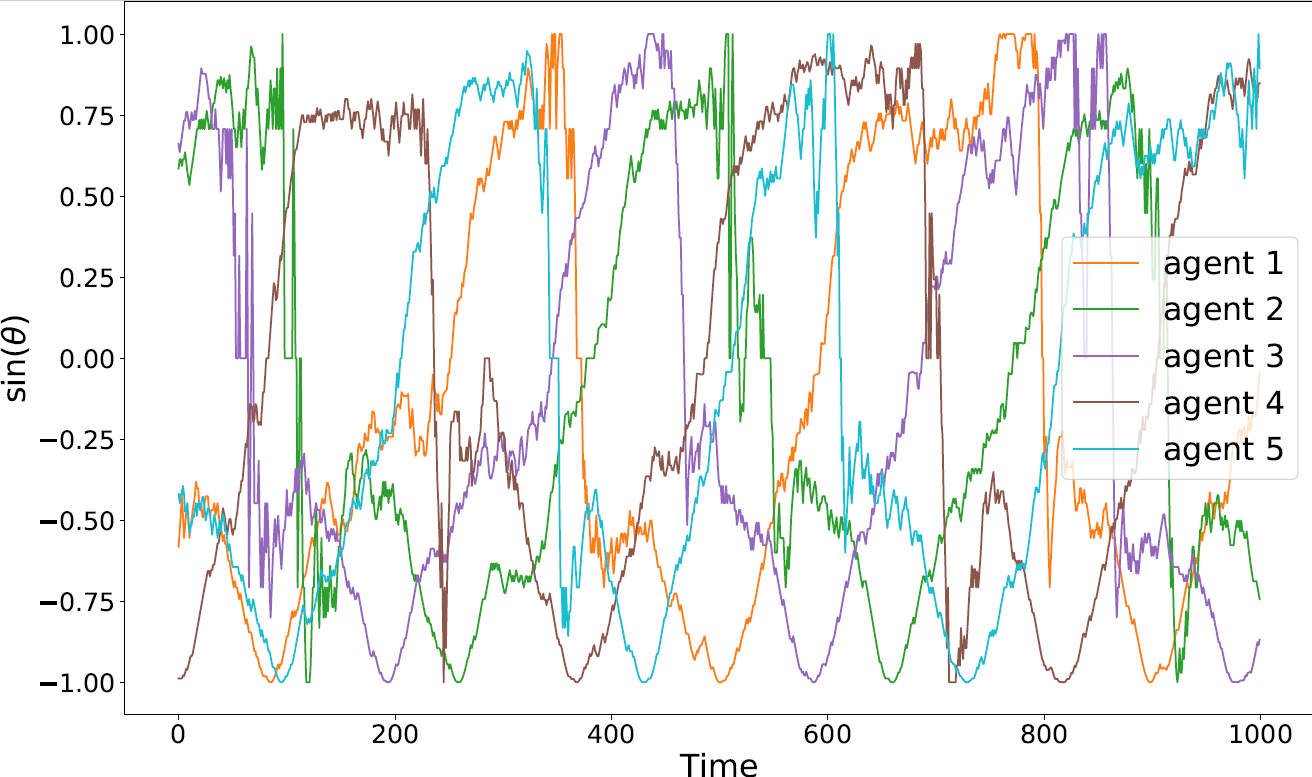}}
\caption{In Open map. (a) Polar-coordinate plot for $N_{test}= 4$. (b) Polar-coordinate plot for deployment $N_{test}= 6$. }
\label{fig:sin_theta_2}
\end{figure}

In contrast, when we use MA-PPO to train the model, the policy we get is as Figure \ref{fig:optimal:open map_b2}.
The learned trajectory is that one agent stays in the upper-left corner and the remaining  {$N_{test} - 1$} agents move along a tour at the lower-right corner. The polar-coordinate plot is shown in Figure \ref{fig:sin_theta_2}.

For MA-PPO, the structure (especially the phase difference part) in the solution of ($N_{train}= 4$,$N_{test}= 4$) does not generalize to ($N_{train}= 4$,$N_{test}= 6$) case. This suggests a lack of emergent behavior in the case of MA-PPO as compared to MA-G-PPO which precludes generalization.

We conjecture that the communication is likely the reason for learning the structure in the solution which leads to MA-G-PPO outperforming MA-PPO.

\section{Conclusion} 
Our emphasis in this paper was to investigate a series of questions regarding the efficacy of using MARL for VPM. The experiment results show that given sufficient training time, the proposed approach may be better than TSPC baseline if there exists a nearly-static solution and GCS baseline otherwise. We also show that agents that use a combination of local and global information far outperform agents trained with only one type of information. Furthermore, the agent trained with higher-resolution local information eventually performs better than those with lower-resolution global information. We also see that communication in MA-G-PPO leads to better performance than no communication. One interesting feature we discover is that MA-G-PPO learns emergent behavior for agents. Specifically, we observe that the agents learn a periodic policy with a fixed phase difference in the agents' paths. As mentioned earlier, we conjecture that MA-G-PPO is able to learn the kernels~\cite{alamdari2014persistent} of infinite horizon paths in some structured environments. Furthermore, this structure is preserved even when the number of agents that are deployed is not the same as the number of agents during training. This leads to better generalization. The results presented here provide evidence of this in Visibility based Persistent Monitoring settings. 

\bibliographystyle{IEEEtran}
\bibliography{IEEEabrv,main,refs}

\end{document}